\documentclass[11pt]{article}

\usepackage[final]{acl}

\usepackage{times}
\usepackage{latexsym}

\usepackage[T1]{fontenc}

\usepackage[utf8]{inputenc}

\usepackage{microtype}

\usepackage{inconsolata}

\usepackage{epigraph}
\usepackage{graphicx}
\usepackage{booktabs}      
\usepackage{multirow}      
\usepackage{arydshln}      
\usepackage{amsmath}
\usepackage{amsmath, amssymb}
\usepackage[table]{xcolor}

\usepackage{float}
\usepackage{placeins}

\usepackage{listings}
\usepackage{xcolor}
\usepackage{tcolorbox}
\tcbuselibrary{listings, breakable, skins}

\definecolor{promptbg}{RGB}{240, 253, 250}
\definecolor{promptframe}{RGB}{45, 180, 160}
\definecolor{titlebg}{RGB}{17, 148, 130}
\definecolor{keywordcolor}{RGB}{8, 126, 139}
\definecolor{stringcolor}{RGB}{180, 100, 50}
\definecolor{commentcolor}{RGB}{115, 130, 125}

\newtcblisting{promptbox}[2][]{
    enhanced,
    colback=promptbg,
    colframe=promptframe,
    colbacktitle=titlebg,
    coltitle=white,
    fonttitle=\bfseries\small,
    title={#2},
    boxrule=0.6pt,
    arc=3pt,
    left=8pt,
    right=8pt,
    top=4pt,
    bottom=4pt,
    toptitle=3pt,
    bottomtitle=3pt,
    listing only,
    listing options={
        basicstyle=\ttfamily\scriptsize\linespread{1.1}\selectfont,
        breaklines=true,
        breakatwhitespace=false,
        columns=fullflexible,
        keepspaces=true,
        showstringspaces=false,
        commentstyle=\color{commentcolor}\itshape,
        keywordstyle=\color{keywordcolor},
        stringstyle=\color{stringcolor},
        xleftmargin=0pt,
        xrightmargin=0pt,
        inputencoding=utf8,
        extendedchars=true,
        literate=
            {•}{{$\bullet$}}1
            {–}{{--}}1
            {—}{{---}}1
            {"}{{``}}1
            {"}{{''}}1
            {'}{{'}}1
            {'}{{'}}1
            {…}{{...}}1
            {→}{{$\rightarrow$}}1
            {←}{{$\leftarrow$}}1
            {≥}{{$\geq$}}1
            {≤}{{$\leq$}}1
            {≠}{{$\neq$}}1,
    },
    #1
}

\title{Glance-or-Gaze: Incentivizing LMMs to Adaptively Focus Search via Reinforcement Learning}

\newcommand{\ourmodel}{GoG}

\author{
\textbf{Hongbo Bai}$^{1*}$, \textbf{Yujin Zhou}$^{1*}$, \textbf{Yile Wu}$^{1}$, \textbf{Chi-Min Chan}$^1$ \\
\textbf{Pengcheng Wen}$^1$, \textbf{Kunhao Pan}$^1$, \textbf{Sirui Han}$^{1\dag}$, \textbf{Yike Guo}$^{1\dag}$\\
$^1$Hong Kong University of Science and Technology \\
\texttt{baihongbo@ust.hk} \\
\texttt{yzhouha@connect.ust.hk}
}

\begin{document}
\maketitle
{
\renewcommand{\thefootnote}{\fnsymbol{footnote}}
\footnotetext[1]{Equal contribution. $^\dag$Corresponding author.}
}

\begin{abstract}

Large Multimodal Models (LMMs) have achieved remarkable success in visual understanding, yet they struggle with knowledge-intensive queries involving long-tail entities or evolving information due to static parametric knowledge. Recent search-augmented approaches attempt to address this limitation, but existing methods rely on indiscriminate whole-image retrieval that introduces substantial visual redundancy and noise, and lack deep iterative reflection, limiting their effectiveness on complex visual queries. To overcome these challenges, we propose Glance-or-Gaze (GoG), a fully autonomous framework that shifts from passive perception to active visual planning. GoG introduces a Selective Gaze mechanism that dynamically chooses whether to glance at global context or gaze into high-value regions, filtering irrelevant information before retrieval. We design a dual-stage training strategy: Reflective GoG Behavior Alignment via supervised fine-tuning instills the fundamental GoG paradigm, while Complexity-Adaptive Reinforcement Learning further enhances the model's capability to handle complex queries through iterative reasoning. Experiments across six benchmarks demonstrate state-of-the-art performance. Ablation studies confirm that both Selective Gaze and complexity-adaptive RL are essential for effective visual search. We will release our code and models for further exploration soon.\footnote{\url{https://github.com/TOM-ZHOUch/Glance-or-Gaze}}

\end{abstract}

\epigraph{\textit{``You see, but you do not observe. The distinction is clear.''}}{--- Arthur Conan Doyle, \textit{A Scandal in Bohemia}}

\section{Introduction}\label{sec:introduction}

\begin{figure*}[t]
    \centering
    \includegraphics[width=1.0\linewidth]{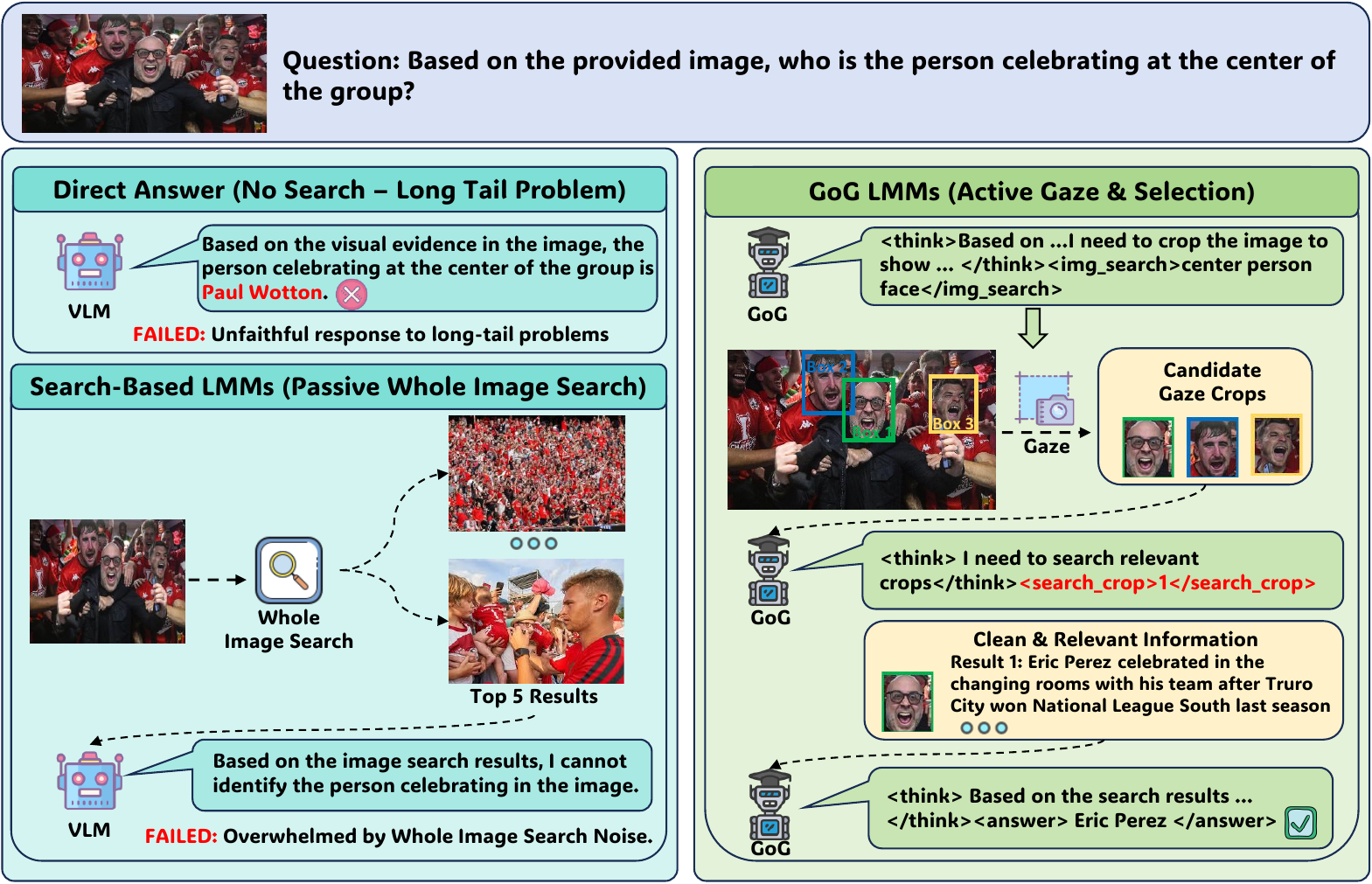}
    \caption{Comparison between previous baselines and our Glance-or-Gaze (GoG) framework. GoG employs an active multi-step strategy: proposing candidate regions via visual grounding, filtering relevant crops through Selective Gaze, and conducting precise search only on selected regions with iterative cross-modal reflection.}
    \label{fig:main}
\end{figure*}

Large Multimodal Models (LMMs) have demonstrated remarkable proficiency in general visual understanding and reasoning, driven by extensive visual-text pre-training~\citep{wang2025scaling,wang2024comprehensive,qi2025vision}. Despite these progress, a fundamental tension persists between the static nature of parametric knowledge and the dynamic, open-ended evolution of the real world. Constrained by fixed training corpora, LMMs inevitably struggle with knowledge-intensive Visual Question Answering (VQA) tasks that require up-to-date information or involve concepts of long-tail beyond their internal distribution~\citep{narayan2025deepmmsearch,li2023comprehensive}. Specifically, when querying recent events or obscure visual entities, such as identifying a niche festival or interpreting a newly released product, these models frequently exhibit ``knowledge cutoff'' or ``long-tail amnesia''~\cite{lewis2020retrieval,hu2023reveal,xu2024search,tong2024eyes}. Such deficiencies often precipitate factual hallucinations or generic responses, thereby undermining the LMMs' utility in knowledge-critical domains~\citep{wei2024uniir,yasunaga2022retrieval,luo2025finmme,ji2025mirage}.

To mitigate this gap, recent research has increasingly sought to augment LMMs with external search mechanisms. Existing works generally fall into three distinct categories. Initially, approaches adopt Retrieval-Augmented Generation (RAG), which retrieves auxiliary context from fixed knowledge bases~\citep{wu2025visual,lin2024mm,lin2023fine}. However, this "retrieve-then-generate" pipeline often suffers from rigidity, failing to capture the dynamic breadth of the open web. Subsequently, Prompt-Engineered Agents emerged, leveraging the planning capabilities of LMMs to orchestrate search engines~\citep{li2025dyfo,wu2024v}. While flexible, these systems typically operate in a "plug-and-play" manner, leaving the model's intrinsic search capabilities unoptimized. More recently, tool-integrated LMMs have advanced this frontier by incorporating search actions directly into the training process, thereby aligning the model's internal representations with external retrieval behaviors~\citep{wu2025mmsearch,narayan2025deepmmsearch}.

Despite recent progress, existing methodologies face two critical limitations. First, current visual strategies are inefficient and noisy. By relying on rigid cropping and indiscriminately processing all candidate regions, these methods introduce significant noise. Moreover, they depend heavily on converting visual details into text descriptions, which creates an information bottleneck and prevents the model from accessing raw visual data. Second, current works execute tool calls in a single-pass manner or limit reflection strictly to the textual modality. They fail to support iterative visual verification or self-correction, restricting their ability to adaptively refine strategies when facing intricate visual questions.

To overcome these limitations, we propose \textbf{Glance-or-Gaze (GoG)}, the first fully autonomous framework that shifts the paradigm from passive image perception to dynamic, complexity-adaptive visual planning, as illustrated in Figure~\ref{fig:main}. We introduce a Selective Gaze mechanism that actively filters visual noise by evaluating and prioritizing pertinent image patches before execution. The framework integrates two complementary learning mechanisms: \textbf{(i) Reflective GoG Behavior Alignment} addresses the inefficiency and noise inherent in current visual strategies by employing imitation learning to instill the fundamental paradigm of active selection and cross-modal reflection. \textbf{(ii) Complexity-Adaptive Reinforcement Learning} further enhances the model's planning and reasoning capabilities by optimizing the selection policy, enabling intelligent, iterative reflection and adaptive focus search based on visual query complexity. Our main contributions are summarized as follows:

\begin{itemize} 
    \item We propose \ourmodel, the first fully autonomous framework that shifts the paradigm from passive image perception to dynamic, complexity-adaptive visual planning. We introduce a Selective Gaze mechanism that actively filters visual noise by evaluating and prioritizing pertinent image patches, effectively bridging the gap between coarse-grained glancing and fine-grained evidence verification.
    
    \item We design a dual-stage learning architecture comprising Reflective GoG Behavior Alignment and Complexity-Adaptive Reinforcement Learning. The former addresses the inefficiency inherent in current visual strategies by instilling the fundamental paradigm of active selection and cross-modal reflection. The latter further enhances planning and reasoning capabilities, enabling adaptive search based on visual query complexity.
    
    \item Extensive experiments demonstrate the effectiveness and generalization of \ourmodel. Our method achieves state-of-the-art performance across diverse benchmarks, surpassing strong baselines with significant gains ranging from 5 to 20, validating the superiority of our adaptive GoG planning paradigm.
\end{itemize}

\section{Related Works}\label{sec:related_works}
\subsection{Large Multimodal Models}

The evolution of Large Multimodal Models (LMMs) has progressed rapidly from early vision-language pretraining~\citep{agarwal2021evaluating,li2022blip,zhu2025safemtmultiturnsafetymultimodal} to sophisticated unified architectures capable of complex reasoning~\citep{liang2026vizomem, bai2025qwen2,yakun2026mmfctub,yakun2026perceptionunderstandingreasoningmultimodal,zhou2026whether}. Recent models such as GPT-4o~\citep{openai2024gpt4ocard}, Gemini~\citep{gemmateam2025gemma3technicalreport}, Qwen3-VL series~\citep{bai2025qwen3vltechnicalreport}, and Internvl3.5-VL~\citep{wang2025internvl3} have demonstrated remarkable capabilities in visual understanding, achieving strong performance on benchmarks spanning image captioning and visual question answering. Despite these advances, LMMs remain fundamentally constrained by their static parametric knowledge~\citep{lewis2020retrieval, hu2023reveal, liu2025datasage}. When confronted with queries involving recent events, long-tail entities, or fine-grained visual details beyond their training distribution, these models frequently hallucinate or produce generic responses~\citep{xu2024search}.

\subsection{Search-Augmented Visual Reasoning}
To address the knowledge limitations of LMMs, researchers have explored various strategies for integrating external information retrieval. Early efforts extended text-based Retrieval-Augmented Generation (RAG) to the multimodal domain, retrieving relevant documents or images from fixed knowledge bases\citep{wu2025visual, lin2024mm, lin2023fine}. Subsequently, prompt-based search agents emerged, leveraging the planning capabilities of LMMs to orchestrate web search engines via chain-of-thought prompting~\citep{yao2022react,yang2023mm,li2025dyfo,wu2024v}. More recently, tool-integrated LMMs have incorporated search actions directly into the training process, with MMSearch-R1~\citep{wu2025mmsearch} embedding search within the reinforcement learning loop and optimizing web navigation policies through online interaction~\citep{wu2025mmsearch, narayan2025deepmmsearch}. Despite these advances, current visual strategies remain inefficient and noisy, relying on rigid cropping and indiscriminate region processing while converting visual details into text, creating an information bottleneck that prevents access to raw visual data. Furthermore, existing approaches execute tool calls in a single-pass manner without iterative visual verification, limiting their ability to adaptively refine strategies for complex visual queries. To address these challenges, we propose Glance-or-Gaze (GoG), a unified framework that integrates active visual exploration with reflective reasoning, enabling dynamic, complexity-adaptive visual planning.

\section{Method}\label{sec:method}

\subsection{Overview}

\begin{figure*}[t]
    \centering
    \includegraphics[width=1.0\linewidth]{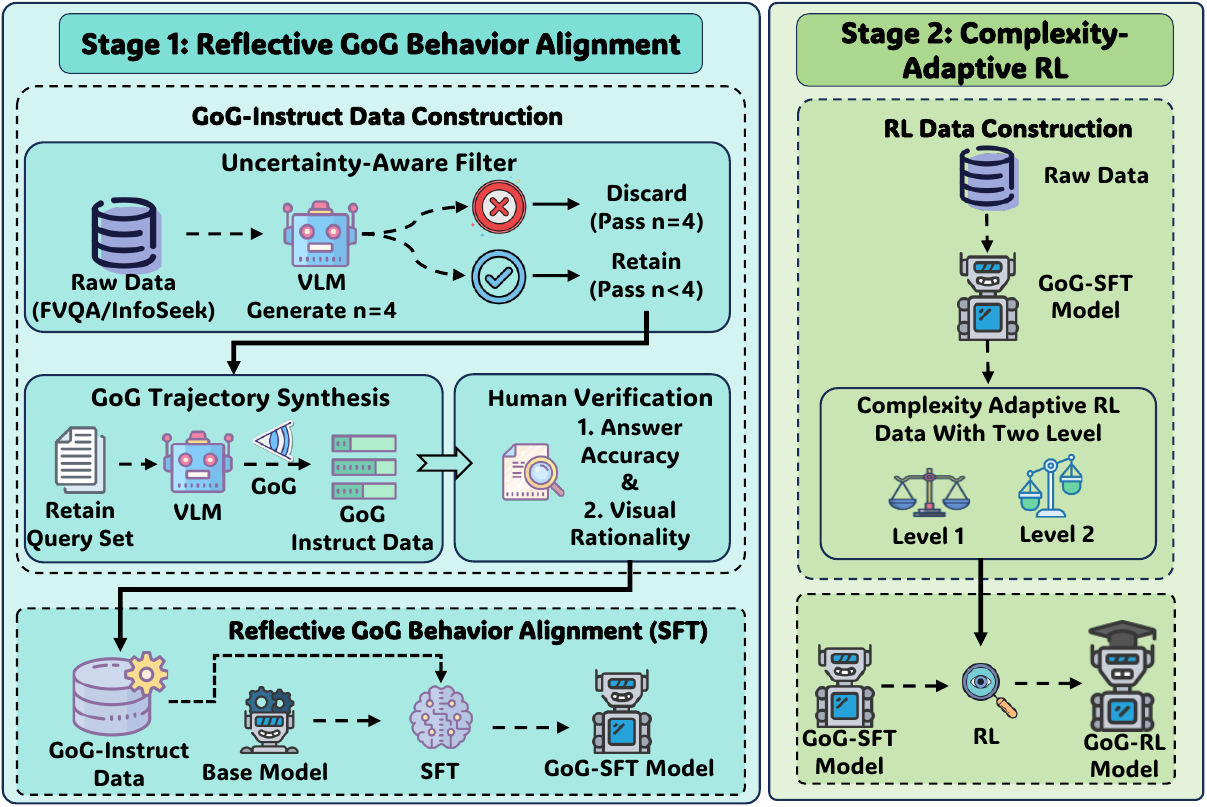}
    \caption{Overview of the Glance-or-Gaze (GoG) framework. \textbf{Stage 1 (Left):} Reflective GoG Behavior Alignment constructs GoG-Instruct data through uncertainty-aware filtering and human-verified trajectory synthesis, then performs supervised fine-tuning to instill active selection and cross-modal reflection. \textbf{Stage 2 (Right):} Complexity-Adaptive RL constructs complexity-stratified data at two difficulty levels and applies reinforcement learning to enhance planning capabilities for adaptive visual reasoning.}
    \label{fig:method}
\end{figure*}

We propose a dual-stage learning framework designed to evolve Large Multimodal Models (LMMs) from passive observers into active visual planners. As illustrated in Figure~\ref{fig:method}, our approach integrates \textit{Reflective GoG Behavior Alignment} and \textit{Complexity-Adaptive Reinforcement Learning} to achieve complexity-adaptive visual reasoning.

\begin{itemize}
    \item \textbf{Stage 1: Reflective GoG Behavior Alignment.} This stage addresses the inefficiency and noise inherent in current visual strategies. We establish a specialized data curation pipeline with uncertainty-aware query sampling and human-verified trajectory synthesis to construct the \textbf{GoG-Instruct} dataset. Through supervised fine-tuning, we instill the fundamental paradigm of active selection and cross-modal reflection, training the model to suppress ungrounded generation and proactively trigger ``Glance'' or ``Gaze'' actions.
    
    \item \textbf{Stage 2: Complexity-Adaptive Reinforcement Learning.} This stage further enhances the model's planning and reasoning capabilities. We construct a complexity-stratified dataset based on failure rates and employ Group Relative Policy Optimization (GRPO)~\citep{shao2024deepseekmath}. This empowers the model to autonomously plan and execute multi-step exploratory searches, enabling adaptive search based on visual query complexity.
\end{itemize}

\subsection{Stage 1: Reflective GoG Behavior Alignment}

This stage instills the fundamental paradigm of active selection and cross-modal reflection through imitation learning.

\paragraph{GoG-Instruct Data Construction}
To facilitate active visual planning, We curate \textit{GoG-Instruct}, a dataset derived from FVQA~\citep{wu2025mmsearch} and InfoSeek~\citep{chen2023can}, through a three-step pipeline as illustrated in Figure~\ref{fig:main} (Left).

\noindent\textbf{1. Uncertainty-Aware Filtering.} We filter out trivial samples where the model already possesses sufficient parametric knowledge. Specifically, we employ Qwen3-VL-235B-Instruct to generate answers for each query $N=4$ times. Samples where the model consistently answers correctly (pass count $n=4$) are discarded, retaining only queries with $n < 4$ that require external visual verification.

\noindent\textbf{2. GoG Trajectory Synthesis.} For the retained query set, we synthesize reasoning trajectories with explicit visual planning structure. Each trajectory follows the sequence: (1) \textit{Glance} for global image analysis, (2) \textit{Decision} for tool invocation or region cropping, and (3) \textit{Gaze} for targeted search execution on selected regions.

\noindent\textbf{3. Human Verification.} To ensure data quality, expert annotators validate each trajectory against two criteria: (1) \textbf{Answer Accuracy}, ensuring the final response is factually correct; and (2) \textbf{Visual Rationality}, verifying that cropped regions are logically relevant to the query. Trajectories failing either criterion are discarded.

Through this pipeline, we construct the GoG-Instruct dataset comprising 5,750 samples, where 43.5\% are search-free instances and 56.5\% require various search operations, as detailed in Table~\ref{tab:sft_data}.

\begin{table}[t]
\centering
\small
\begin{tabular}{llcc}
\toprule
\textbf{Type} & \textbf{Subtype} & \textbf{Ratio} & \textbf{Count} \\
\midrule
Search-free & -- & 43.5\% & 2,500 \\
\midrule
\multirow{3}{*}{Search-required} & Text Search Only & 13.0\% & 750 \\
& Image Search Only & 30.4\% & 1,750 \\
& Both Search & 13.0\% & 750 \\
\midrule
\textbf{Total} & & 100\% & 5,750 \\
\bottomrule
\end{tabular}
\caption{Composition of the SFT training dataset. ``Image Search Only'' refers to instances that execute either a single whole image search operation or a Gaze operation (cropped region search), but not both.}
\label{tab:sft_data}
\end{table}

\paragraph{Training Objective}
We employ Qwen2.5-VL-7B-Instruct and Qwen3-VL-Think as the base model. To enable efficient training, we incorporate LoRA adapters ($r=8$) across all transformer blocks. The model is trained on multi-turn conversations $y^*$ containing reasoning sequences and structured tool calls. We adopt the standard Causal Language Modeling (Causal LM) objective:
\begin{equation}
\mathcal{L}_{\text{SFT}} = -\sum_{t=1}^{T} \log \pi_\theta(y^*_t \mid x, I, y^*_{<t}),
\end{equation}
where $T$ is the sequence length and $\pi_\theta$ is the model's conditional distribution.

\subsection{Stage 2: Complexity-Adaptive Reinforcement Learning}

While SFT establishes the capability for active planning, it lacks the flexibility to handle varying query difficulties. We employ Reinforcement Learning to optimize the planning policy for robustness, as illustrated in Figure~\ref{fig:main} (Right).

\paragraph{RL Data Construction}
We construct complexity-stratified data by evaluating the GoG-SFT model on raw queries and stratifying them into two difficulty levels:

\noindent\textbf{Level 1:} This subset consists of queries where the GoG-SFT model exhibits a pass rate of approximately 50\% using standard ``Glance'' or ``Gaze'' actions. These samples represent the decision boundary where the model is uncertain, providing effective signal for policy optimization.

\noindent\textbf{Level 2:} Building upon Level 1, this subset incorporates additional queries where the GoG paradigm consistently fails, resulting in an overall lower pass rate. Training on this level forces the model to develop more robust reasoning strategies beyond the standard visual exploration patterns.

We adopt Level 2 data as our final RL training set to maximize the model's capability for complex visual reasoning.

\paragraph{Training Objective}
We employ GRPO to optimize the policy on the Level 2 dataset $\mathcal{D}_{\text{L2}}$. For each multimodal input $(q, I) \in \mathcal{D}_{\text{L2}}$, we sample a group of $G$ candidate trajectories $\{\tau_i\}_{i=1}^G$ from the current policy $\pi_\theta$. Each trajectory receives a reward $r_i$, and we compute the group-normalized advantage $\hat{A}_i = (r_i - \mu_r) / \sigma_r$, where $\mu_r$ and $\sigma_r$ are the mean and standard deviation of rewards within the group. The objective function is:
\begin{equation}
\small
\mathcal{L}_{\text{GRPO}}(\theta) = \mathbb{E}_{(q, I) \sim \mathcal{D}_{\text{L2}}} \left[ \frac{1}{G} \sum_{i=1}^G \mathcal{L}_i^{\text{clip}} - \beta \mathbb{D}_{\text{KL}}(\pi_\theta \| \pi_{\text{ref}}) \right]
\end{equation}
where $\mathcal{L}_i^{\text{clip}}$ is the clipped surrogate objective and $\beta$ controls the KL divergence penalty to prevent excessive deviation from the reference model $\pi_{\text{ref}}$.

The reward $r_i$ combines accuracy and format compliance. The accuracy score $r_{\text{acc}} \in \{0, 1\}$ is judged by \texttt{gpt-oss-120b} based on semantic correctness against ground truth. The format score $r_{\text{fmt}} \in [0, 1]$ validates correct usage of special tokens and structural integrity. The total reward is:
\begin{equation}
r_i = (1 - \lambda) \cdot r_{\text{acc}} + \lambda \cdot r_{\text{fmt}}
\end{equation}
where $\lambda$ controls the emphasis on formatting compliance.

\section{Experiments}\label{sec:exp}
\subsection{Experiment Setup}
\paragraph{Implementation Details} We initialize our backbone using Qwen2.5-VL-7B-Instruct and Qwen3-VL-8B-Think. For SFT, we utilize the LLaMA-Factory framework with LoRA (rank 8) applied to all target modules. The models are trained for 3 epochs with a learning rate of $1e^{-4}$, a cosine learning rate scheduler, and a warmup ratio of 0.1. We employ bf16 mixed precision with a global batch size of 8. For online RL optimization, we adopt the Group Relative Policy Optimization (GRPO) algorithm within the veRL framework. We set the actor learning rate to $2e^{-6}$ with a sigmoid decay warmup over 45 steps. The KL-divergence coefficient is set to $\beta = 0.001$. During exploration, we use a rollout number of $N=4$ with vLLM for efficient inference, allowing a maximum response length of 8,192 tokens. The RL training runs for 15 epochs. All training are conducted on a single node equipped with 8 NVIDIA H800 GPUs. Additional hyperparameter details are provided in Appendix~\ref{app:implementation}.

\paragraph{Benchmark and Metrics} Benchmarks. To comprehensively evaluate \ourmodel, we conduct experiments on a diverse set of datasets categorized into in-distribution (IID) and out-of-distribution (OOD) settings. For IID evaluation, we utilize InfoSeek~\citep{chen2023can} and FVQA-test~\citep{wu2025mmsearch}. To assess the model's generalization capability (OOD), we employ SimpleVQA~\citep{cheng2025simplevqamultimodalfactualityevaluation}, MMSearch~\citep{jiang2024mmsearchbenchmarkingpotentiallarge}, DynVQA~\citep{li2024benchmarking}, and LiveVQA-New~\citep{fu2025seekingupdatinglivevisual}. Due to the substantial size of the original splits, we randomly sample 2,000 instances from InfoSeek and LiveVQA for evaluation. Additionally, we filter all datasets to exclusively retain English-language samples. Following prior work, we employ an LLM-as-a-Judge approach to evaluate model performance. Specifically, we utilize \texttt{gpt-oss-120b} as the judge model to directly assess the accuracy of the generated answers. The full evaluation prompt used for this assessment is provided in Appendix~\ref{app:prompts}.

\paragraph{Baselines.} To benchmark the capabilities of \ourmodel, we evaluate it against a series of strong baselines, including both open-source models (the Qwen2.5-VL and Qwen3-VL series) and closed-source models (GPT-4o~\citep{openai2024gpt4ocard}). We primarily compare performance across four distinct settings: (1) \textbf{Direct Answer}, where the model is provided with the image and question and asked to respond directly; (2) \textbf{Full-Search Workflow}, where retrieval is mandatory for every query; (3) \textbf{Prompt-based GoG Agents}, which utilize Graph-of-Thought reasoning via prompting; and (4) \textbf{Search-Equipped LMMs}, specifically comparing against MMSearch R1. Detailed evaluation protocols and the specific prompts used are provided in Appendix~\ref{app:prompts}.

\begin{table*}[!t]
\centering
\renewcommand{\arraystretch}{1.1}
\setlength{\tabcolsep}{3.5pt}
\small
\definecolor{ourscolor}{RGB}{230, 243, 255}
\definecolor{baselinecolor}{RGB}{245, 245, 245}

\begin{tabular}{l c | c c | c c c c}
\toprule
& & \multicolumn{2}{c|}{\textit{In-Domain}} & \multicolumn{4}{c}{\textit{Out-of-Domain}} \\
\textbf{Model} & \textbf{Avg.} & \textbf{FVQA} & \textbf{InfoSeek} & \textbf{SimpleVQA} & \textbf{MMSearch} & \textbf{LiveVQA} & \textbf{DynVQA} \\
\midrule
\multicolumn{8}{c}{\textit{Direct Answer}} \\
\cdashline{1-8}
Qwen2.5-VL-7B              & 20.31 & 25.33 & 13.90 & 41.26 & 14.62 & 10.30 & 16.43 \\
Qwen2.5-VL-32B             & 22.79 & 26.89 & 16.95 & 42.84 & 16.37 & 12.70 & 20.96 \\
Qwen3-VL-8B-Instruct       & 21.13 & 23.94 & 13.80 & 40.18 & 15.20 & 11.40 & 22.24 \\
Qwen3-VL-8B-Thinking       & 23.84 & 24.56 & 16.05 & 41.76 & 15.20 & 15.15 & 30.31 \\
Qwen3-VL-30B-A3B-Instruct  & 23.80 & 26.11 & 16.70 & 42.35 & 18.13 & 12.60 & 26.91 \\
Qwen3-VL-30B-A3B-Thinking  & 27.26 & 31.11 & 22.80 & 43.73 & 17.54 & 15.40 & 33.00 \\
GPT-4o                     & 30.68 & 42.00 & 30.60 & 43.44 & 21.64 & 14.80 & 31.59 \\
\midrule
\multicolumn{8}{c}{\textit{Prompt-based GoG Method}} \\
\cdashline{1-8}
Qwen2.5-VL-7B   & 22.31 & 24.50 & 17.40 & 41.66 & 20.47 & 10.40 & 19.41 \\
Qwen3-VL-8B-Think & 41.70 & 51.33 & 32.00 & 62.69 & 36.84 & 25.95 & 41.36 \\
\midrule
\multicolumn{8}{c}{\textit{Full-Search Workflow}} \\
\cdashline{1-8}
Qwen2.5-VL-7B & 44.36 & 61.06 & 38.15 & 59.13 & 36.26 & 31.05 & 40.51 \\
Qwen3-VL-8B-Thinking   & 46.99 & 57.33 & 32.25 & 61.90 & 63.84 & 23.55 & 39.09 \\
\midrule
\multicolumn{8}{c}{\textit{Search-Equipped Models}} \\
\cdashline{1-8}
\rowcolor{baselinecolor} MMSearch-R1*                    & 36.91 & 42.39 & 24.65 & 54.79 & 40.94 & 23.85 & 34.84 \\
\rowcolor{ourscolor} \ourmodel-2.5-7B-SFT              & 43.28 & 53.72 & 41.90 & 60.61 & 40.35 & 24.55 & 38.53 \\
\rowcolor{ourscolor} \ourmodel-3-8B-Think-SFT          & 50.17 & 62.17 & 40.55 & 65.65 & 53.80 & 32.40 & 46.46 \\
\rowcolor{ourscolor} \ourmodel-2.5-7B-RL           & 53.22 & 66.78 & \textbf{51.05} & 64.86 & 56.73 & 37.70 & 42.21 \\
\rowcolor{ourscolor} \ourmodel-3-8B-Think-RL       & \textbf{56.88} & \textbf{68.44} & 49.05 & \textbf{66.44} & \textbf{65.50} & \textbf{43.85} & \textbf{48.02} \\
\bottomrule
\end{tabular}
\caption{Main evaluation results. We compare Direct Answer, Prompt-based GoG, Full-Search Workflow, and Search-Equipped Models. \colorbox{ourscolor}{Blue rows} indicate our proposed models (SFT and RL). \colorbox{baselinecolor}{Gray row} indicates the reproduced MMSearch-R1 baseline. * denotes reproduced results.}
\label{tab:main_results}
\end{table*}
\subsection{Main Results and Observations}

\noindent{\textbf{\ourmodel~achieves state-of-the-art performance with robust generalization across diverse benchmarks.}} As detailed in Table~\ref{tab:main_results}, \ourmodel-3-8B-Think-RL surpasses all baseline paradigms, outperforming the strongest Full-Search Workflow and Prompt-based GoG agent by +9.89 and +15.18 on average, respectively. Notably, compared to the previous search-equipped baseline MMSearch-R1, \ourmodel~achieves a remarkable +19.97 improvement. This superiority is consistent across both in-domain datasets and out-of-domain settings, demonstrating the model's exceptional adaptability. We observe that while Full-Search workflows improve over Direct Answer baselines (such as GPT-4o) by retrieving external knowledge, they often fall short of ~\ourmodel because they lack the learned discrimination to filter irrelevant noise or decide when search is strictly necessary. 

\begin{figure}[t]
    \centering
    \includegraphics[width=1.0\linewidth]{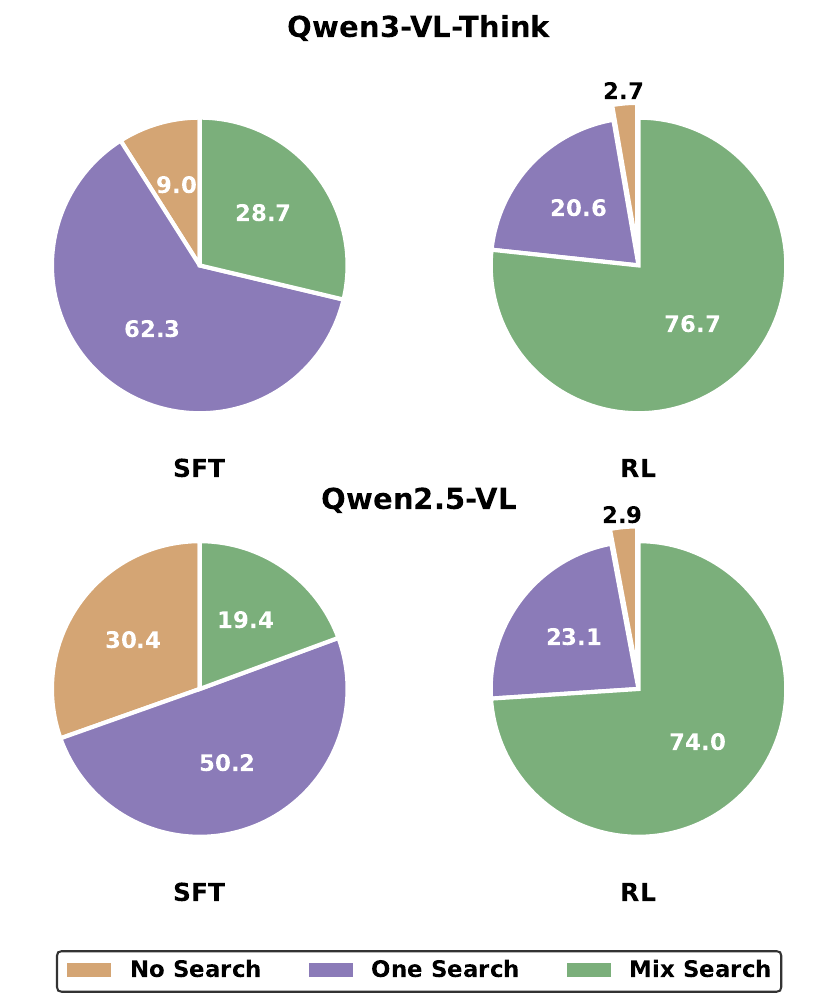}
    \caption{Distribution of search behavior across different training stages. ``No Search'' indicates samples without any search action, ``One Search'' represents samples using only one type of search (text, image, or crop), and ``Mix Search'' denotes samples combining multiple search types.}
    \label{fig:search_distribution}
\end{figure}

\noindent{\textbf{SFT teaches basic GoG capability while RL enhances interative GoG reasoning.}}
As shown in Figure~\ref{fig:search_distribution}, SFT and RL exhibit distinct search behavior patterns. After SFT, both models primarily rely on single-type searches (62.3\% for Qwen3-VL-Think and 50.2\% for Qwen2.5-VL), with a notable proportion of samples requiring no search at all (9.0\% and 30.4\%, respectively). This suggests that SFT successfully teaches the model \textit{when} and \textit{how} to invoke search tools for straightforward queries. However, after RL training, the proportion of mix search increases dramatically---from 28.7\% to 76.7\% for Qwen3-VL-Think and from 19.4\% to 74.0\% for Qwen2.5-VL. Meanwhile, the no-search ratio drops to below 3\% for both models. This shift indicates that RL encourages the model to engage in more complex, multi-step information gathering, combining text search, image search, and image cropping to thoroughly verify and cross-reference information before answering. The emergence of such iterative search behavior demonstrates that RL not only reinforces correct search usage but also cultivates a more deliberate reasoning process.

\begin{figure}[t]
    \centering
    \includegraphics[width=1.0\linewidth]{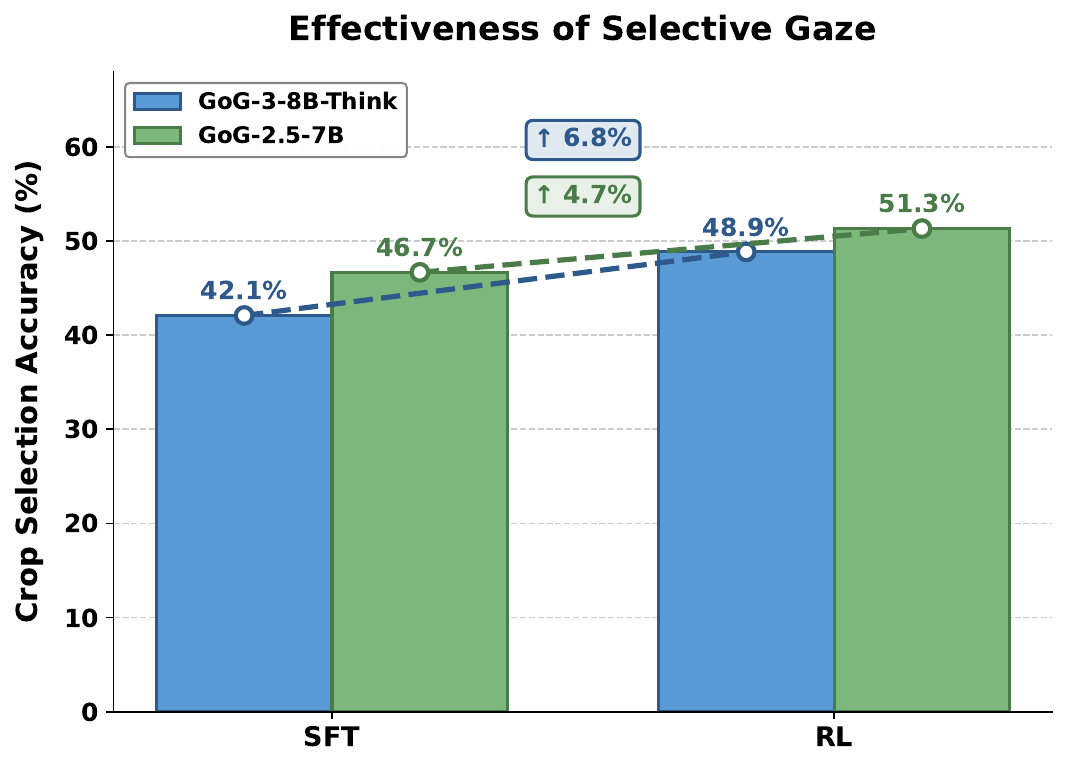}
    \vspace{-2mm} 
    \caption{\textbf{Effectiveness of Selective Gaze.} Comparison of Crop Selection Accuracy between SFT and RL training stages across two model architectures.}
    \label{fig:crop_effectiveness}
    \vspace{-3mm} 
\end{figure}

\noindent\textbf{RL training sharpens Selective Gaze to identify answer-relevant regions more precisely.} Beyond overall performance gains, we investigate whether RL training specifically improves the model's ability to select relevant image regions. As shown in Figure~\ref{fig:crop_effectiveness}, we measure \textit{Crop Selection Accuracy}—the proportion of selected crops that contain answer-relevant information—on our evaluation set. Both models exhibit substantial improvements after RL training: Qwen3-VL-Think improves from 42.1\% to 48.9\% (\textbf{+6.8\%}), while Qwen2.5-VL increases from 46.7\% to 51.3\% (\textbf{+4.7\%}). This consistent gain across different architectures demonstrates that RL does not merely encourage more frequent tool use, but fundamentally refines \textit{where} the model chooses to focus. To gain deeper insights, we manually examined 100 samples from GoG-3-8B-Think where Gaze operations were invoked (Table~\ref{tab:gaze_analysis}). We evaluated two aspects: (1) \textit{Gaze Correctness}—whether the selected crop is appropriate for answering the question, and (2) \textit{Reflection Rate}—among incorrect selections, whether the model triggers self-correction. Results show that RL training improves Gaze correctness from 59\% to 75\%, and more strikingly, enhances error awareness: when the initial Gaze is incorrect, SFT models rarely recognize the error (30\% reflection rate), whereas RL-trained models trigger reflection 70\% of the time. This indicates that RL instills better-calibrated uncertainty, enabling the model to recognize misguided focus and actively seek alternative regions.

\begin{table}[t]
\centering
\small
\label{tab:gaze_analysis}
\begin{tabular}{lcc}
\toprule
\textbf{Metric} & \textbf{SFT} & \textbf{RL} \\
\midrule
Gaze Correctness (\%) & 59 & 75 \\
Reflection Rate on Errors (\%) & 30 & 70 \\
\bottomrule
\end{tabular}
\caption{Manual analysis of Gaze behavior on 100 GoG-3-8B-Think samples.}
\label{tab:gaze_analysis}
\end{table}

\begin{table}[t]
\centering
\renewcommand{\arraystretch}{1.1}
\setlength{\tabcolsep}{3.5pt} 
\resizebox{\linewidth}{!}{
\begin{tabular}{l c | c c c c c c}
\toprule
\textbf{Model Setting} & \textbf{Avg.} & \textbf{FVQA} & \textbf{Info} & \textbf{Simple} & \textbf{MM} & \textbf{Live} & \textbf{Dyn} \\
\midrule
\multicolumn{8}{l}{\textit{Base Model: Qwen2.5-VL-7B}} \\
\rowcolor{blue!5}
\quad SFT w/o SG & 41.76 & 53.39 & 40.10 & 59.53 & \textbf{40.94} & 22.30 & 34.28 \\
\rowcolor{blue!10}
\quad \textbf{Full SFT} & \textbf{43.28} & \textbf{53.72} & \textbf{41.90} & \textbf{60.61} & 40.35 & \textbf{24.55} & \textbf{38.53} \\
\rowcolor{green!8}
\quad \textit{$\Delta$ Gain} & \cellcolor{green!15}\textit{+1.52} & \textit{+0.33} & \textit{+1.80} & \textit{+1.08} & \cellcolor{red!10}\textit{-0.59} & \textit{+2.25} & \cellcolor{green!20}\textit{+4.25} \\
\midrule
\multicolumn{8}{l}{\textit{Base Model: Qwen3-VL-8B-Thinking}} \\
\rowcolor{blue!5}
\quad SFT w/o SG & 48.34 & 60.17 & 40.55 & 64.86 & 46.20 & \textbf{32.80} & 45.47 \\
\rowcolor{blue!10}
\quad \textbf{Full SFT} & \textbf{50.17} & \textbf{62.17} & 40.55 & \textbf{65.65} & \textbf{53.80} & 32.40 & \textbf{46.46} \\
\rowcolor{green!8}
\quad \textit{$\Delta$ Gain} & \cellcolor{green!15}\textit{+1.83} & \textit{+2.00} & \textit{+0.00} & \textit{+0.79} & \cellcolor{green!25}\textit{+7.60} & \cellcolor{red!10}\textit{-0.40} & \textit{+0.99} \\
\bottomrule
\end{tabular}
}
\caption{Ablation study results. We compare the full SFT model against the ablated version (SFT w/o SG), where \textbf{SG} denotes the \textbf{Selective Gaze mechanism}. \colorbox{green!20}{Green} highlights notable gains; \colorbox{red!10}{red} indicates slight regressions.}
\label{tab:ablation_results}
\end{table}
\subsection{Ablation and Analysis}

\noindent\textbf{Effectiveness of the Selective Gaze Mechanism.} 
To validate our architectural design, we conduct an ablation study by removing the Selective Gaze (SG) mechanism. In this "w/o SG" variant, the model loses the ability to perform dynamic visual cropping and reflection, regressing to a baseline that relies solely on a coarse-grained global view of retrieved images. As shown in Table~\ref{tab:ablation_results}, this removal leads to consistent performance degradation across both Qwen2.5 and Qwen3 backbones, with average scores dropping by 1.52 and 1.83 points, respectively. The results underscore the necessity of the "Selective-Gaze-and-Reflect" paradigm, particularly in complex scenarios like DynVQA (+4.25 gain) and MMSearch (+7.60 gain) where target information is often minute or obscured. The SG mechanism acts as a physical anchor for reasoning: by explicitly directing its gaze to a selected candidate region, it forces the model to transition from passively \textit{seeing} the global context to actively \textit{observing} specific visual details. This focused verification provides the granular evidence necessary to ground the model's reasoning, preventing hallucinations common in noise-heavy environments.

\begin{table}[t]
\centering
\renewcommand{\arraystretch}{1.1}
\setlength{\tabcolsep}{3.5pt} 
\resizebox{\linewidth}{!}{
\begin{tabular}{l c | c c c c c c}
\toprule
\textbf{Model Setting} & \textbf{Avg.} & \textbf{FVQA} & \textbf{Info} & \textbf{Simple} & \textbf{MM} & \textbf{Live} & \textbf{Dyn} \\
\midrule
\multicolumn{8}{l}{\textit{Base Model: Qwen2.5-VL-7B}} \\
\rowcolor{blue!5}
\quad RL w/ Leval 1 Data & 47.28 & 64.00 & 50.35 & 64.66 & 50.88 & 32.95 & 39.94 \\
\rowcolor{blue!10}
\quad \textbf{RL w/ Leval 2 Data} & \textbf{53.22} & \textbf{66.78} & \textbf{51.05} & \textbf{64.86} & \textbf{56.73} & \textbf{37.70} & \textbf{42.21} \\
\rowcolor{green!8}
\quad \textit{$\Delta$ Gain} & \cellcolor{green!25}\textit{+5.94} & \textit{+2.78} & \textit{+0.70} & \textit{+0.20} & \cellcolor{green!20}\textit{+5.85} & \cellcolor{green!18}\textit{+4.75} & \textit{+2.27} \\
\midrule
\multicolumn{8}{l}{\textit{Base Model: Qwen3-VL-8B-Thinking}} \\
\rowcolor{blue!5}
\quad RL w/ Level 1 Data & 48.89 & 66.39 & 44.95 & 64.86 & 61.99 & 39.55 & 45.61 \\
\rowcolor{blue!10}
\quad \textbf{RL w/ Level 2 Data} & \textbf{52.38} & \textbf{68.44} & \textbf{49.05} & \textbf{66.44} & \textbf{65.50} & \textbf{43.85} & \textbf{48.02} \\
\rowcolor{green!8}
\quad \textit{$\Delta$ Gain} & \cellcolor{green!20}\textit{+3.49} & \textit{+2.05} & \cellcolor{green!18}\textit{+4.10} & \textit{+1.58} & \textit{+3.51} & \cellcolor{green!18}\textit{+4.30} & \textit{+2.41} \\
\bottomrule
\end{tabular}
}
\caption{Ablation study on Complexity-Adaptive RL data construction. \textbf{Level 2 Data} refers to samples with SFT model pass rate $<50\%$, while \textbf{Level 2 Data} includes samples with pass rate $\geq 50\%$. \colorbox{green!20}{Green} highlights substantial gains ($\geq$4\%). Training on hard samples consistently yields improvements across both backbones.}
\label{tab:complexity_ablation}
\end{table}

\noindent\textbf{Effectiveness of Complexity-Adaptive RL Training.} 
We further investigate the impact of data difficulty on policy optimization, positing that samples challenging for the SFT model provide richer learning signals. To this end, we partition the RL training pool into \textbf{Level 2} (pass rate $<50\%$) and \textbf{Level 1} (pass rate $50\%$). As shown in Table~\ref{tab:complexity_ablation}, training on hard samples consistently outperforms the easy data setting. Specifically, for Qwen2.5-VL-7B, the hard data setting achieves a +5.94 average gain over the easy setting, with significant boosts on benchmarks requiring complex reasoning, such as MMSearch (+5.85) and LiveVQA (+4.75). Similarly, Qwen3-VL-8B-Think demonstrates a +3.49 average improvement with consistent gains across all metrics. These results confirm that while easy samples offer limited room for policy refinement, hard samples present genuine decision-making challenges that maximize the benefits of exploratory optimization, thereby validating our complexity-adaptive data construction strategy.

\section{Conclusion}\label{sec:con}
We present Glance-or-Gaze (GoG), a framework that transforms Large Multimodal Models from passive observers into active visual planners. GoG introduces the Selective Gaze mechanism to dynamically allocate attention between global context and fine-grained regions, filtering visual noise before retrieval. Our dual-stage training strategy—Reflective Behavior Alignment via SFT followed by Complexity-Adaptive Reinforcement Learning—progressively instills cross-modal planning capabilities and refines decision-making on challenging queries. Experiments across six benchmarks demonstrate state-of-the-art performance, with GoG surpassing strong baselines by substantial margins. We believe this work establishes a promising direction for building more capable and autonomous multimodal search agents.

\section{Limitations}\label{sec:limitation}
While GoG demonstrates strong performance across diverse benchmarks, several aspects warrant further investigation. First, although we have optimized the stability of our search infrastructure and Jina Reader pipeline, we observe an occasional failure rate of approximately 1–5\% due to network instability, API timeouts, or malformed webpage content. Such failures may cause incomplete information retrieval and degrade answer quality in affected cases. Second, our experiments primarily focus on English-language benchmarks, and the generalization of GoG to multilingual or cross-lingual visual question answering remains unexplored. Extending the framework to support diverse languages and culturally specific visual knowledge presents an interesting direction for future work.

\section*{Acknowledgments}

This work is funded in part by the HKUST Start-up Fund (R9911), Theme-based Research Scheme grant (T45-205/21-N), the InnoHK initiative of the Innovation and Technology Commission of the Hong Kong Special Administrative Region Government, and the research funding under HKUST-DXM AI for Finance Joint Laboratory (DXM25EG01).

\bibliography{custom}

\appendix

\label{sec:appendix}

\section{Ethics Statement}

We discuss the ethical considerations and licensing of resources used in this work. Our framework is built upon publicly available models and datasets, and we ensure compliance with their respective licenses.

\paragraph{Model Licenses.} We utilize Qwen2.5-VL-7B-Instruct and Qwen3-VL-8B-Think as base models, both released under the Apache-2.0 license, which permits academic research and modification.

\paragraph{Dataset Licenses.} All evaluation benchmarks used in this work are publicly available for research purposes:
\begin{itemize}
    \item \textbf{SimpleVQA}: Apache-2.0 license
    \item \textbf{LiveVQA-new}: CC-BY-NC-4.0 license
    \item \textbf{FVQA}: Apache-2.0 license
    \item \textbf{DynVQA}: Apache-2.0 license
    \item \textbf{MMSearch}: Available for research use
    \item \textbf{InfoSeek}: Apache-2.0 license
\end{itemize}

\paragraph{Human Annotation.} The full annotation workflow, including detailed instructions and quality-control procedures, is described in Section~\ref{sec:method}. All annotators provided informed consent after being briefed on the task objectives, expected workload, and data usage policies. Annotators were compensated at fair market rates for their expertise and time.

\paragraph{Intended Use.} Our GoG framework is designed for research purposes in multimodal understanding and knowledge-intensive visual question answering. We do not anticipate direct negative societal impacts from this work. However, as with any system that retrieves information from the web, users should be aware that retrieved content may contain inaccuracies or biases present in online sources.

\section{Datasets}
\label{sec:appendix_datasets}

We evaluate our method on six diverse benchmarks that assess different aspects of knowledge-intensive visual question answering. Below we provide detailed descriptions of each dataset.

\subsection{FVQA}

FVQA is a factual visual question answering dataset designed to evaluate models on knowledge-requiring visual questions. The dataset is constructed through a combination of automated and manual annotation processes.

The training set, FVQA-train, comprises 5,000 samples with a balanced distribution of approximately 3,400 search-required and 1,600 search-free examples. These samples are collected from three sources: (1) FVQA-auto-vc, containing 5,400 training samples generated by retrieving image-webpage pairs for 10,000 visual concepts sampled from the MetaCLIP metadata distribution, then using GPT-4o to generate factual VQA pairs; (2) FVQA-auto-txt, consisting of 7,000 samples derived from the InfoSeek dataset through balanced sampling across knowledge categories; and (3) FVQA-manual-train, comprising 800 manually annotated samples where annotators selected knowledge categories, located relevant images, and formulated factual questions with precise answers.

The test set, FVQA-test, contains 1,800 high-quality examples that are either manually verified or fully human-annotated. It includes 600 samples from FVQA-auto-vc with manual verification, 600 samples from the InfoSeek Human Split with manually annotated answers, and 600 samples from direct manual annotation.

\subsection{InfoSeek}

InfoSeek is a visual question answering dataset tailored for information-seeking questions that cannot be answered with common sense knowledge alone. The dataset is constructed through a semi-automated process that transforms Wikidata triples into natural language questions using human-authored templates. Annotators design question templates for 300 Wikidata relations, incorporating placeholders for visual entity types and units. These questions are paired with corresponding images and answers to form image-question-answer triplets. To ensure diversity and answerability, question-answer pairs lacking supporting evidence in Wikipedia are filtered out, and balanced sampling is applied across entities and relations. We randomly sampled 2,000 examples from its test split for evaluation.

\subsection{SimpleVQA}

SimpleVQA is a benchmark designed to evaluate factual knowledge boundaries of multimodal large language models. The dataset consists of 2,025 samples spanning 9 core tasks and 9 primary domains. The tasks include Logic \& Science, Object Identification Recognition, Time \& Event, Person \& Emotion, Location \& Building, Text Processing, Quantity \& Position Relationship, Art \& Culture, and Object Attributes Recognition. The domains cover Literature, Education \& Sports, Euro-American History \& Culture, Contemporary Society, Engineering, Technology \& Application, Film, Television \& Media, Natural Science, Art, Chinese History \& Culture, and Life. All samples follow a short-answer format with standardized answers, enabling objective assessment through direct answer matching.

\subsection{MMSearch}

MMSearch is a comprehensive evaluation benchmark for assessing multimodal search performance. The dataset contains 300 manually collected instances spanning 14 subfields, with no overlap with current model training data to ensure that correct answers can only be obtained through searching. The benchmark evaluates models on three individual tasks (requery, rerank, and summarization) and one end-to-end task involving a complete search process.

\subsection{LiveVQA}

LiveVQA is a dataset designed to evaluate how multimodal large language models handle up-to-date visual information beyond their training data cutoff. The dataset features 107,143 samples across 12 categories, drawn from recent news articles, video platforms, and academic publications from April 2024 to May 2025. The benchmark specifically tests models on content that extends beyond their knowledge boundaries and evaluates methods for updating models with new visual knowledge.

\subsection{Dyn-VQA}

Dyn-VQA is a challenging dataset comprising 1,452 dynamic questions that require complex multimodal knowledge retrieval strategies. The dataset includes three types of dynamic questions: (1) questions with rapidly changing answers, where context knowledge updates frequently and retrieved content may mix outdated and newer information; (2) questions requiring multi-modal knowledge, demanding retrieval across diverse modalities with tailored retrieval APIs; and (3) multi-hop questions that necessitate varied reasoning steps for solution. Unlike existing VQA datasets that primarily focus on two-hop questions, Dyn-VQA requires flexible planning of retrieval queries, tools, and timing.

\section{Tools}
\label{sec:appendix_tools}

This section describes the tools integrated into our framework. We incorporate two categories of tools: search tools (text search and image search) and a grounding tool. The search tools enable the model to retrieve external knowledge from the web, while the grounding tool allows fine-grained visual entity localization before searching.

\subsection{Search Tools}

\paragraph{Image Search Tool.} Our image search tool is built upon SerpAPI. Given an input image URL, SerpAPI returns a set of visually similar webpages along with metadata including URLs, thumbnails, and titles. We rank the returned results by relevance and extract up to five valid results, each represented as a thumbnail-title pair. In our experiments, all input images and grounded visual entities are uploaded to an image hosting service and mapped to corresponding URLs before being passed to the model.

\paragraph{Text Search Tool.} The text search tool comprises a complete pipeline consisting of three components: SerpAPI-based text search, Jina Reader, and a summarizer. When the model generates a text search query, it is first sent to SerpAPI, which returns the top-5 retrieval results. Jina Reader then processes the URLs and converts the webpage content into structured text. Finally, we employ Qwen3-32B as a summarizer to extract only the information relevant to the question, filtering out irrelevant content. The entire pipeline is executed in parallel, with Qwen3-32B deployed on 8$\times$H800 GPUs.

\subsection{Grounding Tool}

Our grounding tool is based on Grounding DINO. When the model outputs an image search query, the query is passed to our encapsulated Grounding DINO service deployed on 8$\times$H800 GPUs. Grounding DINO returns the top-$n$ bounding boxes most similar to the query, where $n=5$ in our implementation. Through manual inspection, we observed that Grounding DINO often returns redundant results with overlapping regions. To address this issue, we propose a Gaze Selection mechanism that selects the 1 to $n$ most question-relevant grounding results. The selected regions are then sent in parallel to the image search tool, and the retrieved information about the grounded content is returned to the model for further reasoning.

\section{Implementation Details}\label{app:implementation}

This section provides comprehensive implementation details for our training and evaluation pipeline, covering three stages: supervised fine-tuning (SFT), reinforcement learning (RL), and evaluation.

\begin{table}[t]
\centering
\small
\begin{tabular}{l|c}
\toprule
\textbf{Hyperparameter} & \textbf{Value} \\
\midrule
\multicolumn{2}{c}{\textit{Supervised Fine-Tuning (SFT)}} \\
\midrule
Optimizer & AdamW \\
Learning Rate & $1 \times 10^{-5}$ \\
LR Scheduler & Cosine \\
Warmup Ratio & 0.1 \\
Batch Size (Global) & 128 \\
Epochs & 3 \\
Max Sequence Length & 32,768 \\
Precision & bf16 \\
DeepSpeed Stage & ZeRO-3 \\
LoRA Rank & 8 \\
LoRA Target & All \\
\midrule
\multicolumn{2}{c}{\textit{Reinforcement Learning (RL)}} \\
\midrule
Algorithm & GRPO \\
Actor Learning Rate & $1 \times 10^{-6}$ \\
Critic Learning Rate & $1 \times 10^{-5}$ \\
KL Penalty in Reward & False \\
Actor KL Loss Coefficient & 0.001 \\
Rollout Sample Size ($N$) & 5 \\
Max Prompt Length & 8,192 \\
Max Response Length & 8,192 \\
Max Generation Rounds & 5 \\
Total Epochs & 15 \\
Global Train Batch Size & 256 \\
Image Search Limit & 3 \\
Text Search Limit & 3 \\
Parallel Tool Call Threads & 4 \\
\bottomrule
\end{tabular}
\caption{Hyperparameters for SFT and RL training stages.}
\label{tab:hyperparameters}
\end{table}

\subsection{SFT Training Setting}

We initialize our model from Qwen2.5-VL-7B-Instruct and Qwen3-VL-Think and perform supervised fine-tuning using the LLaMA-Factory framework. We adopt LoRA~\citep{hu2022lora} with rank 8 applied to all target modules for parameter-efficient training. The maximum image resolution is set to 262,144 pixels (equivalent to $512 \times 512$) and video resolution to 16,384 pixels. We use the AdamW optimizer with a learning rate of $1 \times 10^{-5}$, a cosine learning rate scheduler, and a warmup ratio of 0.1. Training is conducted for 3 epochs with a global batch size of 128, a maximum sequence length of 32,768 tokens, and bf16 mixed precision. We employ DeepSpeed ZeRO-3 for memory-efficient distributed training. The detailed hyperparameters are summarized in Table~\ref{tab:hyperparameters}.

\subsection{RL Training Setting}

For reinforcement learning, we adopt the Group Relative Policy Optimization (GRPO) algorithm within the veRL framework. The actor learning rate is set to $1 \times 10^{-6}$ with sigmoid decay warmup over 45 steps, while the critic learning rate is $1 \times 10^{-5}$. We disable KL penalty in the reward computation but apply an actor KL loss with coefficient 0.001 to prevent excessive deviation from the reference policy. During rollout, we sample $N=5$ responses per prompt using vLLM with a maximum prompt length of 8,192 tokens and maximum response length of 8,192 tokens. The model is allowed up to 5 rounds of multi-turn tool interactions per episode. For search constraints, we limit image search and text search to 3 calls each, with a maximum of 5 crop rounds. Tool calls are executed in parallel with 4 threads to accelerate exploration. Training runs for 15 epochs with a global batch size of 256, and checkpoints are saved every 50 steps. The complete hyperparameter configuration is provided in Table~\ref{tab:hyperparameters}.

\subsection{Evaluation Setting}

During inference, we deploy our trained models on 2$\times$H800 GPUs using vLLM and serve them as online APIs. We use greedy decoding with temperature set to 0 to ensure reproducible results.

For evaluation, we employ an LLM-as-a-Judge approach following prior work~\citep{wu2025mmsearch}. Specifically, we use GPT-OSS-120B deployed on 8$\times$H800 GPUs as the judge model, with temperature set to 0 for deterministic evaluation. The judge model assesses the correctness and quality of generated answers by comparing them against ground-truth references.

\clearpage
\onecolumn

\section{Prompts}
\label{app:prompts}

This section provides the complete prompts used throughout our framework, including prompts for SFT data generation, inference, summarization, and evaluation. All prompts are carefully designed to guide the model through structured reasoning and tool usage.

\subsection{Prompts for SFT Datasets Generation}

We design a multi-turn prompt system for SFT data generation that guides the model through different stages of the search process. The system consists of four specialized prompts: (1) \textbf{Round 1 Prompt} initializes the reasoning process and instructs the model to analyze the image and select an appropriate action; (2) \textbf{After Image Search Prompt} guides the model to process image search results and decide on follow-up actions; (3) \textbf{After Gaze Search Prompt} handles the selection of cropped regions for further searching; and (4) \textbf{After Text Search Prompt} directs the model to synthesize text search results and formulate the final answer.

\begin{figure*}[htbp]
\begin{promptbox}{Round 1 Prompt}
You are an expert visual assistant. Your task is to answer a user's question based on the provided image.

Step 1: Analyze the Image
Carefully examine the image and the user's question: {question}. Identify all recognizable entities, objects, text, and other visual clues.

Step 2: Plan Your Action
Based on your analysis, you must perform one of the following actions. You must include your thinking process inside a <think>...</think> block before choosing an action. 

    • Action 1: Image Search
        If you are not sure about the visual element and need to identify the visual element in the image, you can use one of the following image search methods.
            – Whole image search: Only use this if the question is about the entire scene in general, its location, or the overall context. Output only: <img_search><img></img_search>.
            Note: Do not output <img_search><img></img></img_search>.
            – Cropped search (recommended for targeted questions): Use this when the question focuses on a particular visual element like an object, person, animal, logo, building, etc. Provide a brief description of the element inside the tags. Examples:
                <img_search>the woman wearing a blue jacket</img_search>
                <img_search>the emblem on the front of the car</img_search>

    • Action 2: Use Text Search
        If you can identify the visual element confidently but need more specific information to answer the question, invoke the text search tool. Generate a focused query and output it as <text_search>your search query</text_search>.

    • Action 3: Direct Answer
        If you are confident in your identification and have enough knowledge to respond, provide a clear and concise answer: <answer>Your answer here.</answer>

Remember, search results will be provided to you in subsequent turn. You can analyze the search results and decide your next action. You can perform image search multiple times (e.g., Whole Image Search first, then Cropped Search if needed), and perform multiple text searches to gather relevant information. All search results will be placed inside <information>...</information>.

Here is the image and the question:
Question: {question}
Image:
\end{promptbox}
\end{figure*}
\FloatBarrier

\begin{figure*}[htbp]
\begin{promptbox}{After Image Search Prompt}
You have received information from an image search. Your goal is to use this new information to answer the original question: {question}.

Step 1: Analyze the Results
Analyze the information provided within the <information>...</information> block. Consider what you have learned about the visual element in question and how it relates to the user's query.

Step 2: Decide Your Next Action
Include your reasoning inside a <think>...</think> block, then select one of the following actions:

    - Action 1: Direct Answer
    If the image search results have helped you identify the visual element and you can confidently answer the question with your internal knowledge, provide the final, concise answer inside an <answer>...</answer> tag.

    - Action 2: Use Text Search
    If the image search results have helped you identify the visual element but you need more specific details to answer the question, invoke the text search tool. Formulate a precise query based on the image search results and output it as <text_search>your search query</text_search>. You can use the text search tool multiple times in subsequent turns if needed.

    - Action 3: Search Image Again
    If the previous image search results were not helpful (e.g., the crop was inaccurate, or whole image search missed details), you can search the image again.
        – If you used Whole Image Search and need details, you MUST try Cropped Search: <img_search>description</img_search>
        – If you used Cropped Search and missed, try refining the description: <img_search>refined description</img_search>.
\end{promptbox}
\end{figure*}

\FloatBarrier

\begin{figure*}[htbp]
\begin{promptbox}{After Gaze Search Prompt}
You have requested to crop the image to focus on specific elements. Here are the cropped images based on your request.
Step 1: Analyze the Cropped Images
Review the cropped images provided above. Determine if they clearly show the visual elements you need to answer the question: {question}.
Step 2: Plan Your Next Action
Include your thinking process inside a <think>...</think> block. Then, choose one of the following actions:
    • Action 1: Search Selected Crops
        If one or more cropped images contain the visual element you need, select them for image search. Output the indices of the crops you want to search (1-based) inside <search_crop>...</search_crop> tags. Example: <search_crop>1,3</search_crop> (to search crop 1 and crop 3).
    • Action 2: Answer Directly
        If the cropped images themselves provide enough information to answer the question without further searching, provide the final answer inside <answer>...</answer> tags.
    • Action 3: Search Again (Crop Again)
        If none of the cropped images are satisfactory (e.g., they missed the object or the crop is too loose), you can try to crop again with a different description. Output the new description inside <img_search>...</img_search> tags. Example: <img_search>the license plate on the car</img_search>.
    • Action 4: Search Again (Whole Image)
        If cropped images are insufficient or the question requires global context that was missed, you can revert to whole image search. Output only: <img_search><img></img_search>.
Note: At this stage, you MUST NOT use text search (<text_search>). Focus on image-based operations only. Text search will be available after the image search results are returned.
\end{promptbox}
\end{figure*}
\FloatBarrier

\begin{figure*}[htbp]
\begin{promptbox}{After Text Search Prompt}
You have acquired new data from a text search. Your objective is to assess this information and determine the optimal next move to address the original inquiry: {question}.
Step 1: Assess the Findings Examine the content within the <information>...</information> block. Cross-reference it with what you already know to identify any remaining gaps needed to fully answer the question.
Step 2: Determine Strategy First, articulate your reasoning process inside a <think>...</think> block. Then, select one of the following courses of action:
    Action 1: Provide Answer If the gathered data is sufficient to cover the request, deliver the final, concise response wrapped in an <answer>...</answer> tag.
    Action 2: Refine Search If the current results are useful but incomplete, conduct a follow-up search. Formulate a more targeted or modified query to locate the missing details. Output this new query as <text_search>your refined search query</text_search>.
    Action 3: Stop Search If you have exhausted reasonable search options and the answer remains elusive based on available data, conclude by outputting <answer>Unable to answer due to lack of relevant information</answer>.
\end{promptbox}
\end{figure*}
\FloatBarrier

\subsection{Prompts for Direct Answer}

To establish a baseline for comparison, we first evaluate models using a direct answer approach where the model receives only the image and question without any external knowledge retrieval. This prompt template instructs the model to provide concise answers based solely on the visual information present in the image.

\begin{figure*}[htbp]
\begin{promptbox}{Direct Answer Prompt}
Based on the image, answer the question with as few words as you can. 
Question: {question} Image: image
\end{promptbox}
\end{figure*}

\FloatBarrier

\subsection{Prompts for Full Search Workflow}

Our full search workflow consists of a two-round prompting strategy designed to leverage external knowledge through simulated search engine interactions.

In the first round, the model is presented with the original question, the input image, and five ranked image search results that provide contextual information related to the query. Each search result includes both a webpage thumbnail and its corresponding title. The model is then tasked with formulating an optimized text query that would effectively retrieve the necessary information from a search engine to answer the original question.

The second round prompt provides the model with the text search results obtained from executing the query generated in the first round. Given these retrieved documents along with the original question and image, the model is instructed to synthesize the information and produce a concise final answer.

\begin{figure*}[htbp]
\begin{promptbox}{1st Round Prompt}
You are presented with a question and an image that requires external knowledge to be answered. To aid your understanding, five ranked image search results related to the original image are provided. Each result contains a webpage image and its title.
Question: {question} Image: {image}
Image Search Results:
1. Webpage Image: {image} Webpage Title: {title}
2. Webpage Image: {image} Webpage Title: {title}
...
5. Webpage Image: {image} Webpage Title: {title}
Assume you have access to a search engine (e.g., Google). Based on the question, the original image, and the search results, formulate a targeted text query to retrieve information necessary to answer the question correctly. Optimize your query for search engine algorithms.
Output the text query directly (strictly without quotes or explanation):
\end{promptbox}
\end{figure*}

\FloatBarrier

\begin{figure*}[htbp]
\begin{promptbox}{2nt Round Prompt}
Please read the text search results and answer the question based on the provided image.
Text Search Results: ...
Original question: {question}
Keep your answer as concise as possible.
\end{promptbox}
\end{figure*}

\subsection{Prompts for Summarization}

After retrieving information from text search, we use a summarization prompt to extract only the relevant content from the retrieved webpages. This step is crucial for filtering out noise and reducing the context length before feeding the information back to the model.

\begin{figure*}[htbp]
\begin{promptbox}{Summarization Prompt}
Synthesize the text extracted from Google Lens search results (titles and descriptions) to describe the primary visual content of the original image. Focus on identifying key elements such as figures, objects, settings, or text. Limit your response to 4-5 sentences.
Extracted Text: 
{formatted_results}
\end{promptbox}
\end{figure*}

\FloatBarrier

\subsection{Prompts for LLM as a Judge}

We adopt an LLM-as-a-Judge approach for evaluation, where a separate language model assesses whether the predicted answer is semantically equivalent to the ground-truth answer. The judge is instructed to consider alternate correct answers and focus on semantic equivalence rather than exact string matching.

\begin{figure*}[htbp]
\begin{promptbox}{LLM as a Judge Prompt}
You are an impartial judge evaluating a model's answer for a visual question answering task.

Question: {question}
Ground-Truth Answer(s): {gt_str}
Predicted Answer: {prediction}

IMPORTANT:
- The Ground-Truth Answer(s) may contain alternate correct answers
- The predicted answer is CORRECT if it is semantically equivalent to at least ONE ground-truth answer
- Respond with ONLY one of these two options:
  [CORRECT]
  [INCORRECT]

Your response:
\end{promptbox}
\end{figure*}

\FloatBarrier

\subsection{Prompts for Reward Model}

During reinforcement learning, we use an LLM as reward model to evaluate the quality of generated responses.

\begin{figure*}[htbp]
\begin{promptbox}{LLM as a Reward Model Prompt}
You are an intelligent and impartial expert evaluator for Question Answering tasks. Your specific task is to determine whether the "Predicted Answer" is factually consistent with the "Ground Truth Answer(s)".

# Input Data
- **Question**: {question}
- **Ground Truth Answer(s)**: {ground_truth} (A list of one or more acceptable reference answers).
- **Predicted Answer**: {prediction} (The model's response to evaluate).

# Evaluation Criteria
Please assess the alignment using the following comprehensive rules. If the Predicted Answer matches **ANY** of the Ground Truth Answers according to these rules, the final result is **True**.

1. **Semantic Equivalence (Core Meaning)**
   - **Synonyms & Aliases**: Accept common abbreviations (e.g., "USA" = "United States"), acronyms, and synonyms.
   - **Formatting**: Ignore differences in casing (lower/upper), punctuation, articles ("a", "the"), and spacing.
   - **Typos**: Ignore minor spelling errors if the core meaning is unambiguous.

2. **Numerical & Logical Flexibility**
   - **Units**: Accept equivalent values in different units (e.g., "1.5 kg" = "1500 g", "\$1M" = "1 million dollars").
   - **Rounding**: Accept reasonable rounding deviations (e.g., "3.14" matches "3.14159").
   - **Ranges**: If the prediction is a range that **contains** the specific ground truth value, it is a MATCH (e.g., Prediction "1990-2000" matches Truth "1995").
   - **Dates**: Accept different date formats (e.g., "Jan 1, 2020" = "2020/01/01").

3. **Specificity & Information Containment**
   - **Superset (Extra Info)**: If the prediction contains the correct answer *plus* additional context that does **NOT** contradict the facts, it is CORRECT. (e.g., Truth: "Paris", Prediction: "Paris, the capital of France" -> MATCH).
   - **Subset (More Specific)**: If the prediction is a more precise/specific version of a general ground truth, it is CORRECT.
   - **Lists**: If the answer is a list of items, the order does NOT matter (e.g., "A and B" = "B and A").

4. **Negative Constraints (When to Reject)**
   - **Contradiction**: If the prediction includes the correct entity but makes a claim that directly *contradicts* the question or truth, it is FALSE.
   - **Wrong Entity**: If the core entity is different (e.g., "Mars" vs "Jupiter"), strict rejection.
   - **Hallucination**: If the prediction adds extra details that are factually wrong and change the answer's validity, it is FALSE.

# Output Format
You must output your decision strictly in the following XML format:

<reason>
[Step-by-step reasoning: 1. Identify key entities in Truth and Prediction. 2. Check for semantic/numerical match. 3. Check for contradictions. Explain why it matches or fails.]
</reason>
<judge>
True
</judge>

(Output "True" if it matches, "False" otherwise)
\end{promptbox}
\end{figure*}

\FloatBarrier

\end{document}